\documentclass[sigconf]{acmart}

\usepackage{algorithmic}
\usepackage{graphicx}
\usepackage{textcomp}
\usepackage{xcolor}
\usepackage{url}
\usepackage{listings}
\usepackage{capt-of}

\AtBeginDocument{%
  }

\copyrightyear{2026}
\acmYear{2026}
\setcopyright{cc}
\setcctype{by}
\acmConference[AGENT '26]{International Workshop on Agentic Engineering }{April 12--18, 2026}{Rio de Janeiro, Brazil}
\acmBooktitle{International Workshop on Agentic Engineering (AGENT '26), April 12--18, 2026, Rio de Janeiro, Brazil}
\acmPrice{}
\acmDOI{10.1145/3786167.3788425}
\acmISBN{979-8-4007-2399-5/2026/04}

\begin{document}

\title{AI in Insurance: Adaptive Questionnaires for Improved Risk Profiling}

\author{Diogo Silva}
\affiliation{%
  \institution{Deloitte and \\
Faculty of Engineering,\\
University of Porto\\}
  \city{Porto}
  \country{Portugal}}
\email{diogofilipe2002@live.com}

\author{João Teixeira}
\affiliation{%
  \institution{Deloitte}
  \city{Porto}
  \country{Portugal}}
\email{joaoteixeira@deloitte.pt}

\author{Bruno Lima}
\affiliation{%
  \institution{LIACC,\\
Faculty of Engineering,\\
University of Porto}
  \city{Porto}
  \country{Portugal}}
\email{brunolima@fe.up.pt}


\begin{abstract}
Insurance application processes often rely on lengthy and standardized questionnaires that struggle to capture individual differences. Moreover, insurers must blindly trust users' responses, increasing the chances of fraud. The ARQuest framework introduces a new approach to underwriting by using Large Language Models (LLMs) and alternative data sources to create personalized and adaptive questionnaires. Techniques such as social media image analysis, geographic data categorization, and Retrieval Augmented Generation (RAG) are used to extract meaningful user insights and guide targeted follow-up questions.

A life insurance system integrated into an industry partner mobile app was tested in two experiments. While traditional questionnaires yielded slightly higher accuracy in risk assessment, adaptive versions powered by GPT models required fewer questions and were preferred by users for their more fluid and engaging experience.

ARQuest shows great potential to improve user satisfaction and streamline insurance processes. With further development, this approach may exceed traditional methods regarding risk accuracy and help drive innovation in the insurance industry.
\end{abstract}



\maketitle

\section{Introduction} \label{sec:introduction}

Insurance underwriting — the process of evaluating client risk — is a cornerstone of profitability and long-term business sustainability. Traditional methods, often relying on lengthy, manually filled questionnaires, are prone to human error, fraud, and fail to capture individual nuances \cite{b1}, leading to unfair premiums for both low and high risk clients.

The exponential growth of Artificial Intelligence (AI), coupled with Big Data from digital platforms and social networks, offers a transformative opportunity \cite{b2}. Innovations like Natural Language Processing (NLP) and Large Language Models (LLMs) can enhance communication with customers, while advanced algorithms (neural networks, decision trees) can boost risk prediction \cite{b3}. However, most insurance questionnaires remain static, ignoring valuable contextual information from external sources.

This work introduces a novel approach that uses AI to dynamically generate personalized insurance questionnaires by incorporating external user insights, enabling tailored risk assessment that adapts to each individual, contrasting with conventional one-size-fits-all forms. However, bringing AI-driven adaptive questionnaires into real-world insurance contexts introduces several key challenges:
\begin{itemize}
    \item Multiline scalability: Most current models are optimized for a single insurance domain, making it difficult to adapt to diverse product lines such as life, health, and property insurance.
    \item Data integration: External user insights are often unstructured, context-dependent, and noisy, requiring advanced preprocessing for AI interpretability, while also meeting strict privacy and compliance requirements \cite{b4, b5}.
    \item Fairness: Imperfect underwriting models risk producing biased offers, disproportionately affecting certain customer segments and potentially undermining trust \cite{b6}.
\end{itemize}

This paper addresses these challenges through the design, implementation, and evaluation of ARQuest (Adaptive Risk Questioning), an adaptive questionnaire framework that leverages LLMs and Retrieval Augmented Generation (RAG) to personalize underwriting flows, integrate heterogeneous sources, and ensure fairness and transparency in risk profiling. This work results from a collaboration with a higher education institution and a multinational professional services company\cite{b7}, ensuring scientific relevance and industry applicability.

To guide our study, we formulated the following research questions:
\begin{description}
    \item[\textbf{RQ1}] How can AI generate personalized insurance questionnaires in a clear way?
    \item[\textbf{RQ2}] What concerns arise when integrating external user insights for insurance?
    \item[\textbf{RQ3}] What limitations do current AI models face in capturing individual risk factors?
    \item[\textbf{RQ4}] How do model confidence and explainability affect AI risk assessments?
    \item[\textbf{RQ5}] How can personalized underwriting solutions scale to different types of insurance?
\end{description}

The remainder of this paper is structured as follows. Section \ref{sec:state_of_the_art} reviews related state-of-the-art solutions. Section \ref{sec:arquest} outlines the methodology behind the ARQuest solution. Section \ref{sec:evaluation} presents the evaluation process and results. Finally, Section \ref{sec:conclusion} summarizes findings and discusses future research directions.

\section{State of the Art} \label{sec:state_of_the_art}

The personalization of insurance questionnaires is largely overlooked in current systems. Risk prediction tasks have an underlying set of processes that are relevant for the individualization of the subscription journey \cite{b3}, such as the way insurers collect data about an applicant, the AI models and techniques employed to analyze risk, and all surrounding ethical concerns.

\subsection{Data Integration} \label{sub:state_of_the_art/data_integration}

With Big Data available on a larger scale and data-driven ML models on the rise \cite{b2}, insurance companies do not just rely on forms and reports to understand the risk profile of an individual. Social media, electronic health records, wearable devices, and geographical information are data sources that enhance these systems \cite{b8}. APIs and web scraping methods extract insights, which then go through a transformation process that converts them to a manageable format \cite{b9}. NLP techniques are focused on processing the unstructured natural language found in the data \cite{b10}, while images can be interpreted by advanced models like BLIP \cite{b11}, which generates textual descriptions. However, social media mining is susceptible to nuances regarding the interpretation of a user’s lifestyle \cite{b10}.

\subsection{Personalized Q\&A} \label{sub:state_of_the_art/personalized_qa}

\begin{figure}[!h]
    \centering
    \includegraphics[width=0.48\textwidth]{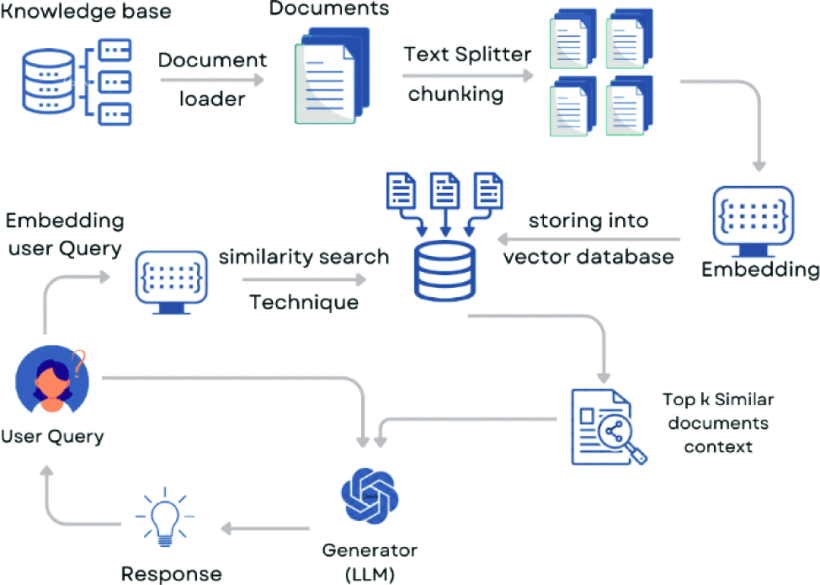}
    \caption{RAG-based question-answering system workflow \cite{b15}.}
    \label{fig:rag}
\end{figure}

LLMs and transformers can process specific datasets to generate personalized questions and answers, which could substitute the preconceived traditional insurance questionnaires. Fine-tuning and prompt-tuning are great modern techniques that leverage these pre-trained general language models and either adapt or contextualize them to the target domain \cite{b12}. Using well-known models like GPT, T5, or BERT variants, different types of questions can be created automatically and then filtered via human evaluation and adequate metrics \cite{b13}. Question answering can be achieved using similar strategies, coupled with semantic similarity for answer quality evaluation \cite{b14}. RAG-based architectures are adopted to mitigate LLM hallucinations in these scenarios, filtering the relevant content that is provided \cite{b15}, as visible in Figure~\ref{fig:rag}.

It is also useful to understand how confident an LLM is in its analysis. The confidence of an answer can be extracted from the internal probabilities of the words generated \cite{b16}.

\subsection{Risk Assessment} \label{sub:state_of_the_art/risk_assessment}

Regarding risk assessment, insurers typically implement a variety of ML algorithms (e.g., random forest, XGBoost, etc.) \cite{b17} and even transformers, which outperform traditional models at interpreting user risk \cite{b18}. The models analyze variables such as age, physical activity, chronic diseases, accident history, and weather conditions to infer if an individual is low or high risk, with additional tasks like fraud detection being executed as well. GPT-4V, for instance, is a powerful model that, following an intelligent prompt, can process the contents of different types of input to capture interesting risk indicators \cite{b19}.

\subsection{Ethical Concerns} \label{sub:state_of_the_art/ethical_concerns}

Due to the insurance implications for user livelihoods, ethical aspects need to be considered when integrating ML models. Compliance with regulatory frameworks like the GDPR \cite{b20} and the AIA \cite{b21} should be a priority, as insurance apps are inherently a high-risk system \cite{b21}. Attempts at protecting consumer data include data encryption and anonymization \cite{b10}. In addition, model discrimination is presented as another vital challenge, as customers from minority groups may be affected by biased premium attribution \cite{b6}. Resampling techniques aid in transforming a dataset so that it becomes representative of the population \cite{b22}. Selecting proper evaluation benchmarks and explainability methods, such as SHAP \cite{b23}, LIME \cite{b23}, or prompting-based techniques \cite{b24}, is also essential to integrate fairness and transparency into the analytical and decision-making process as much as possible.

\section{ARQuest} \label{sec:arquest}

The ARQuest (Adaptive Risk Questioning) framework is presented as a solution to personalize insurance questionnaires by adjusting the questions to the users, aiming at an improved subscription experience and a more precise risk assessment.

\begin{figure}[!h]
    \centering
    \includegraphics[width=0.48\textwidth]{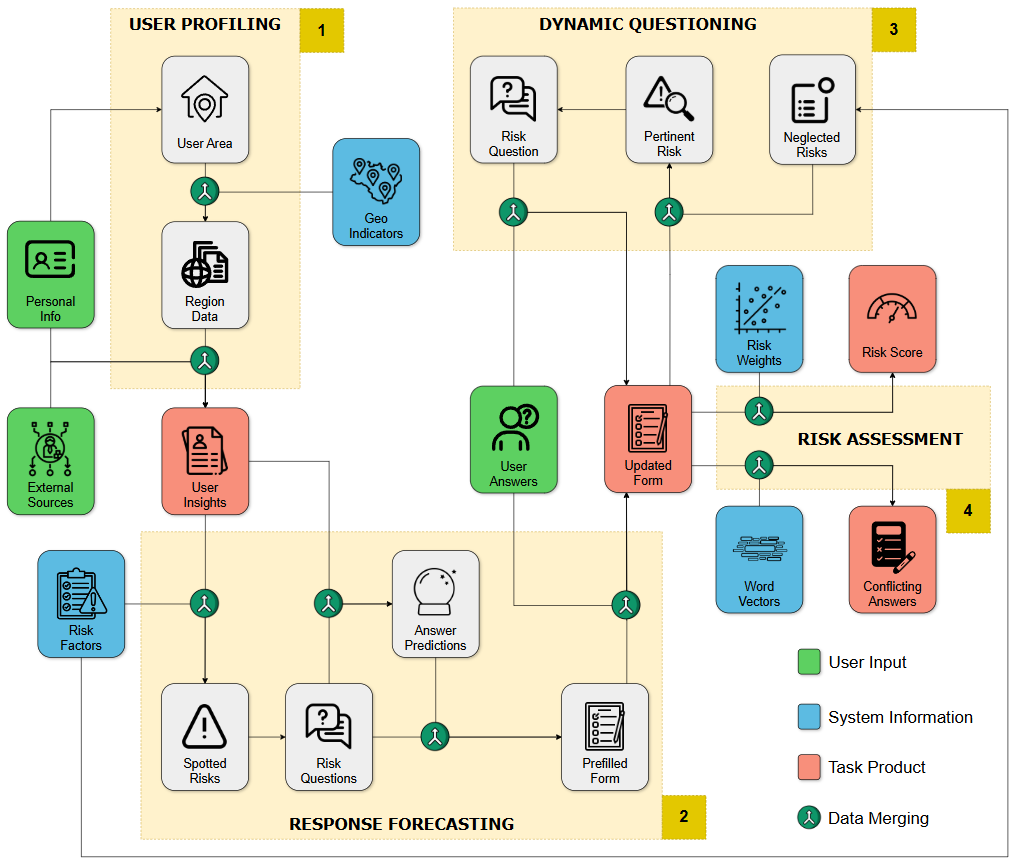}
    \caption{ARQuest framework and its modules.}
    \label{fig:framework}
\end{figure}

The framework, detailed in Figure~\ref{fig:framework}, is divided into four unique modules for different tasks:
\begin{enumerate}
    \item \textbf{User Profiling:} the applicant provides some basic personal information, such as the location, which is used to retrieve relevant geographical risk indicators; data from external platforms can also be shared, with all this information being processed and aggregated.
    \item \textbf{Response Forecasting:} by analyzing the collected insights, an LLM identifies risks from a pre-defined list; these risks have questions assigned, and the LLM predicts the respective answers beforehand; optionally, the user can review and correct these predictions.
    \item \textbf{Dynamic Questioning:} some risks are not detectable from the initial data, so the LLM chooses the next factor (from the remaining) to question the user; this process is repeated, and the user continuously provides answers until the model considers enough information is acquired; briefly, the LLM is prompted to stop the questioning process when none of the remaining factors are both impactful or likely to yield a risky response.
    \item \textbf{Risk Assessment:} based on the completed questionnaire, a global risk score is computed using a specific model or algorithm; cases where the LLM's predictions are drastically corrected by the user are also identified using a similarity measure between guesses and answers; if this occurs frequently, either the LLM performed poorly in its analysis or the applicant intentionally misled the evaluation.
\end{enumerate}

\subsection{System Implementation} \label{sub:arquest/system_implementation}

\begin{figure}[!h]
    \centering
    \includegraphics[width=0.48\textwidth]{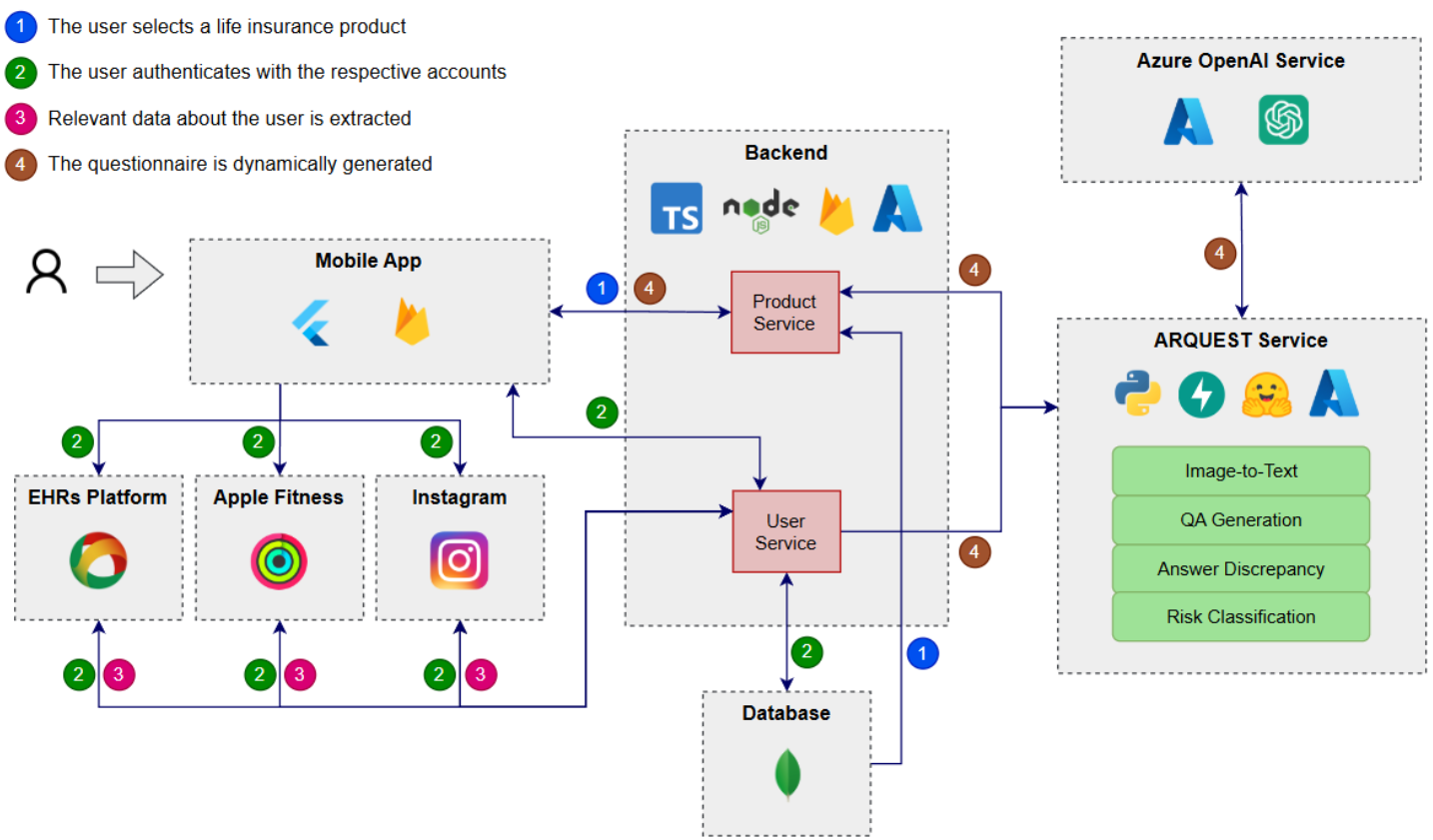}
    \caption{System's flow and architecture.}
    \label{fig:architecture}
\end{figure}

A solution focused on \textbf{life insurance} is implemented within a mobile app project from an industry partner. Figure~\ref{fig:architecture} showcases the overall flow and architecture behind this system.

As depicted, the user can share health records, fitness data, or Instagram posts. The backend communicates with a special service dedicated to the solution's tasks. A GPT model is employed via Azure services \cite{b25} that deal with data protection.

\subsection{Data Preparation} \label{sub:arquest/data_preparation}

\textbf{Life insurance questions} gather information to assess the mortality risk of an individual. Internal and external factors play a role here, such as age, health, genetics, habits, environment, local services, and community. Based on templates from insurance companies [\citenum{b26}, \citenum{b27}], a dataset of multiple-choice questions is created, separated by the following categories: personal details, lifestyle and habits, family history, and health status. The goal is to provide the LLM with a set of risk factors from which it can choose to address, depending on the applicant.

\begin{figure}[!h]
    \centering
    \includegraphics[width=0.48\textwidth]{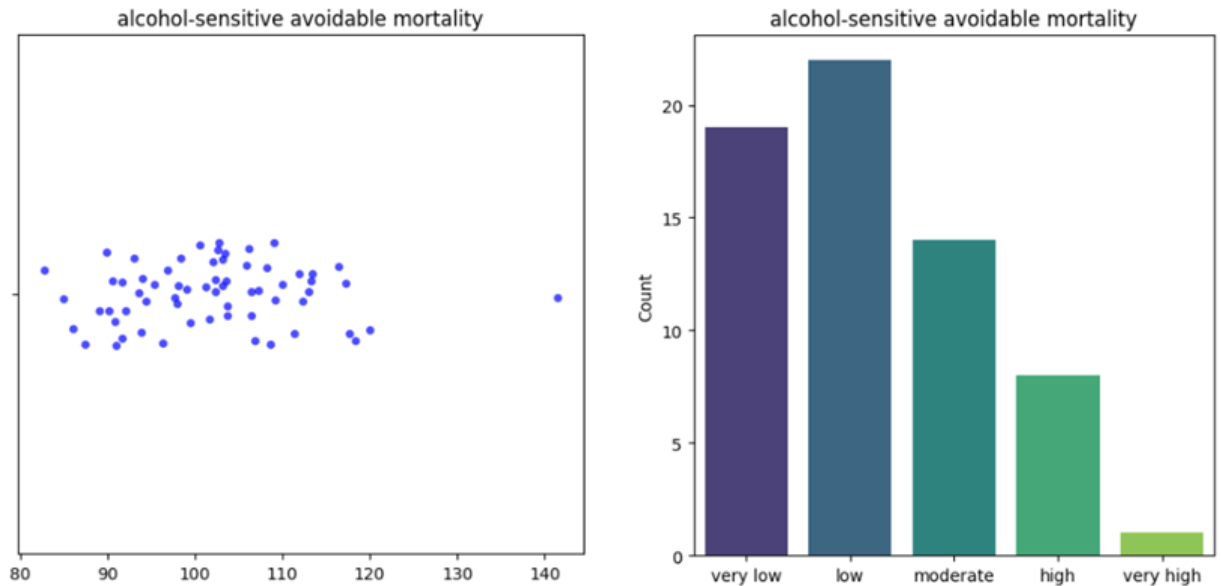}
    \caption{Raw and labeled municipality values for alcohol mortality.}
    \label{fig:clustering}
\end{figure}

The system resorts to \textbf{geographical indicators} collected from the Atlas of Healthy Municipalities website \cite{b28}. It contains statistics on Portugal's municipalities, spread across nine health categories: mortality, morbidity, healthcare, lifestyle, education, socioeconomic status, environment, infrastructure, and security. There are a total of 53 individual priority indicators. The data is extracted via web scraping methods. Additionally, all values are labeled using the k-means clustering algorithm, mentioned by Biddle \textit{et al.} \cite{b29}. In this way, the LLM instantly recognizes that a certain indicator is either "very low" or "very high", instead of having to interpret numerical values with no context surrounding them. Figure~\ref{fig:clustering} demonstrates how raw data points are categorized for a given regional indicator.

The creation of a large set of synthetic users can be helpful in statistically assessing the system's performance. \textbf{Electronic health records} (EHRs) are initially collected from SyntheaTM \cite{b30}, an open-source synthetic patient generation platform with realistic health data. The records contain information on conditions, encounters, medications, procedures, and other relevant aspects.

\begin{figure}[!h]
    \centering
    \includegraphics[width=0.48\textwidth]{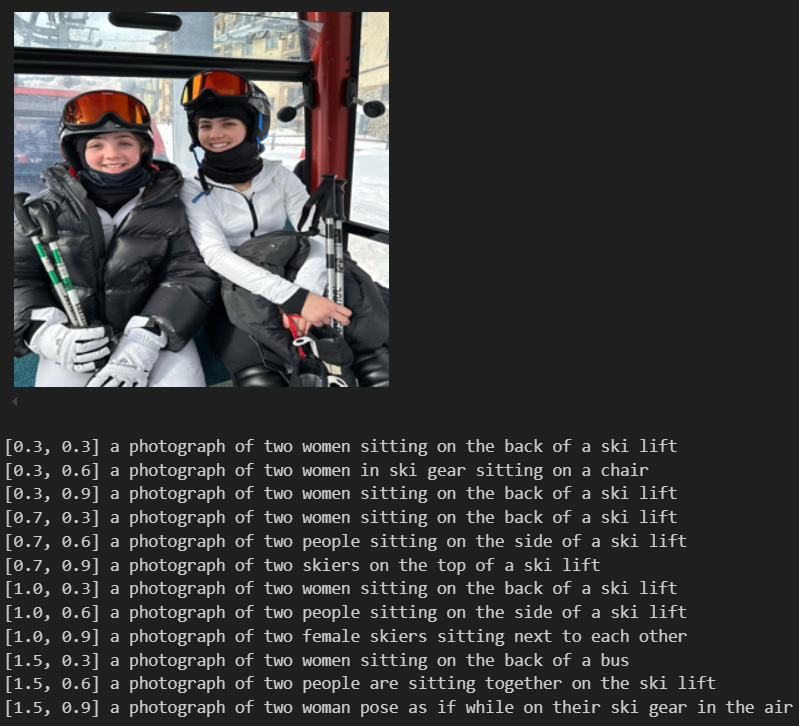}
    \caption{Captions with different parameters for an image \cite{b31}.}
    \label{fig:captioning}
\end{figure}

Social media is useful to understand user interests and activities. Interesting insights, such as risky hobbies, can be captured from this data source. Images from thousands of \textbf{Instagram posts} are gathered from a Kaggle dataset \cite{b31}. Figure~\ref{fig:captioning} displays one of the many experiments that take place with the BLIP model \cite{b11} to comprehend its image captioning capabilities. A specific combination of hyperparameters (\textit{temperature=0.3}, \textit{top-p=0.9}) offers an effective balance between determinism and creativity. This approach is a lightweight and direct solution that semantically enriches the user information fed to the LLM.

The \textbf{synthetic applicant dataset} should strive to represent the target population. The following components are utilized to create a total of 85 artificial users:
\begin{itemize}
    \item A sample of health records matching the real age and gender distributions.
    \item The population across Portugal's municipalities.
    \item A coherent allocation of image descriptions from posts.
    \item Probabilities of working in different occupations.
    \item Average daily steps (fitness metric) estimated using jobs.
    \item Probabilities of sharing external sources.
\end{itemize}

\subsection{Questionnaire Generation} \label{sub:arquest/questionnaire_generation}

For each user, answers to all possible questions from the dataset are pre-computed, resorting to the individual information stored and probabilities. These saved answers act as "ground truths" and automate questionnaire filling upon the execution of tests. Considering a question addressing family diabetes, for instance, if the user's health records mention diabetes and the regional indicators present a high prevalence of the disease, the answer probabilities will favor a positive answer, as it is likely that a family member suffers from that condition, which possesses a hereditary component to it.

\begin{figure}[!h]
    \centering
    \includegraphics[width=0.3\textwidth]{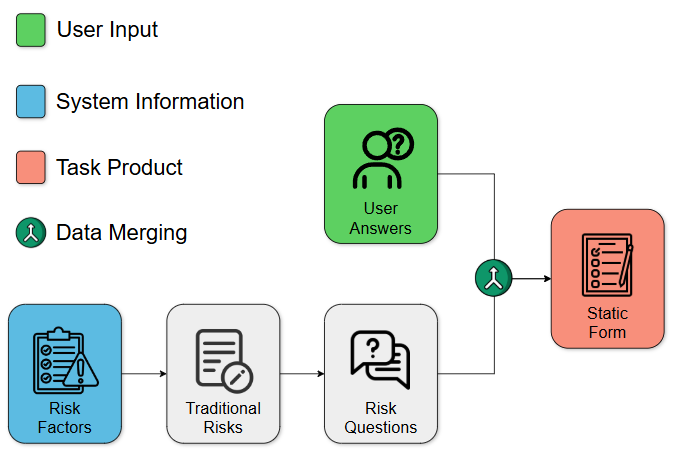}
    \caption{Traditional flow.}
    \label{fig:traditional}
\end{figure}

Two approaches are compared: \textbf{traditional} and \textbf{dynamic}. The traditional questionnaires are static and composed of a pre-defined subset of factors from the question dataset, since real questionnaires do not cover all possible issues, only the most common and influential ones. In this case, they contain 30 questions about lifestyle and habits, family history, and health status, 10 for each category. The described methodology is visible in Figure~\ref{fig:traditional}, where the user is presented with pre-selected questions.

\begin{figure}[!h]
    \centering
    \includegraphics[width=0.48\textwidth]{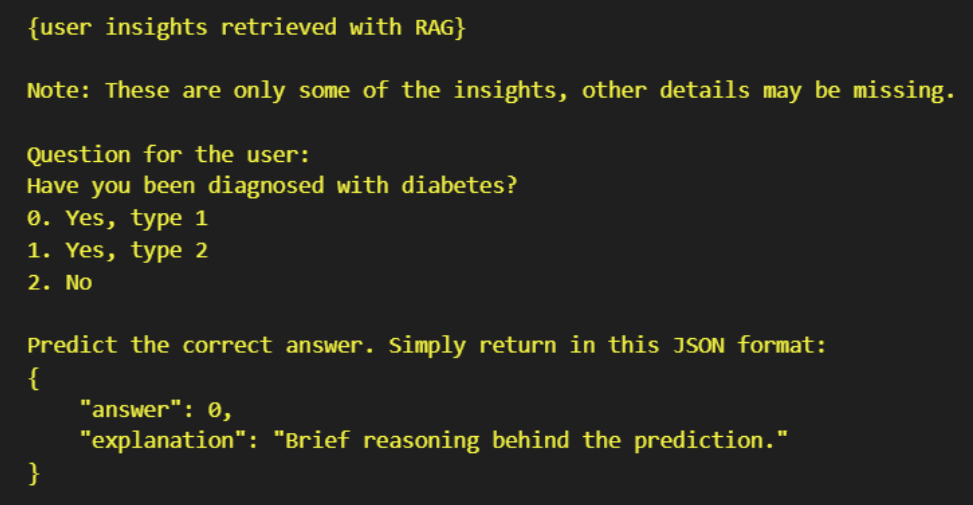}
    \caption{Prompt with answer prediction instructions.}
    \label{fig:prompt}
\end{figure}

The dynamic approach is based on the ARQuest framework, resorting to a GPT model. Figure~\ref{fig:prompt} showcases the prompt given to the model containing answer prediction instructions, which is crucial in the response forecasting task. User insights are initially included in the input. They are gathered using the RAG methodology \cite{b15}, which performs a similarity search by querying all risk factors from the dataset against all available data in the user's external sources. The intended return format (JSON) is presented at the end. The prediction confidence can be extracted by analyzing the token probabilities within the answer field value. An explanation is demanded to reinforce transparency and enhance reasoning.

\subsection{Mobile App Integration} \label{sub:arquest/mobile_app_integration}

\begin{figure}[!h]
    \centering
    \includegraphics[width=0.48\textwidth]{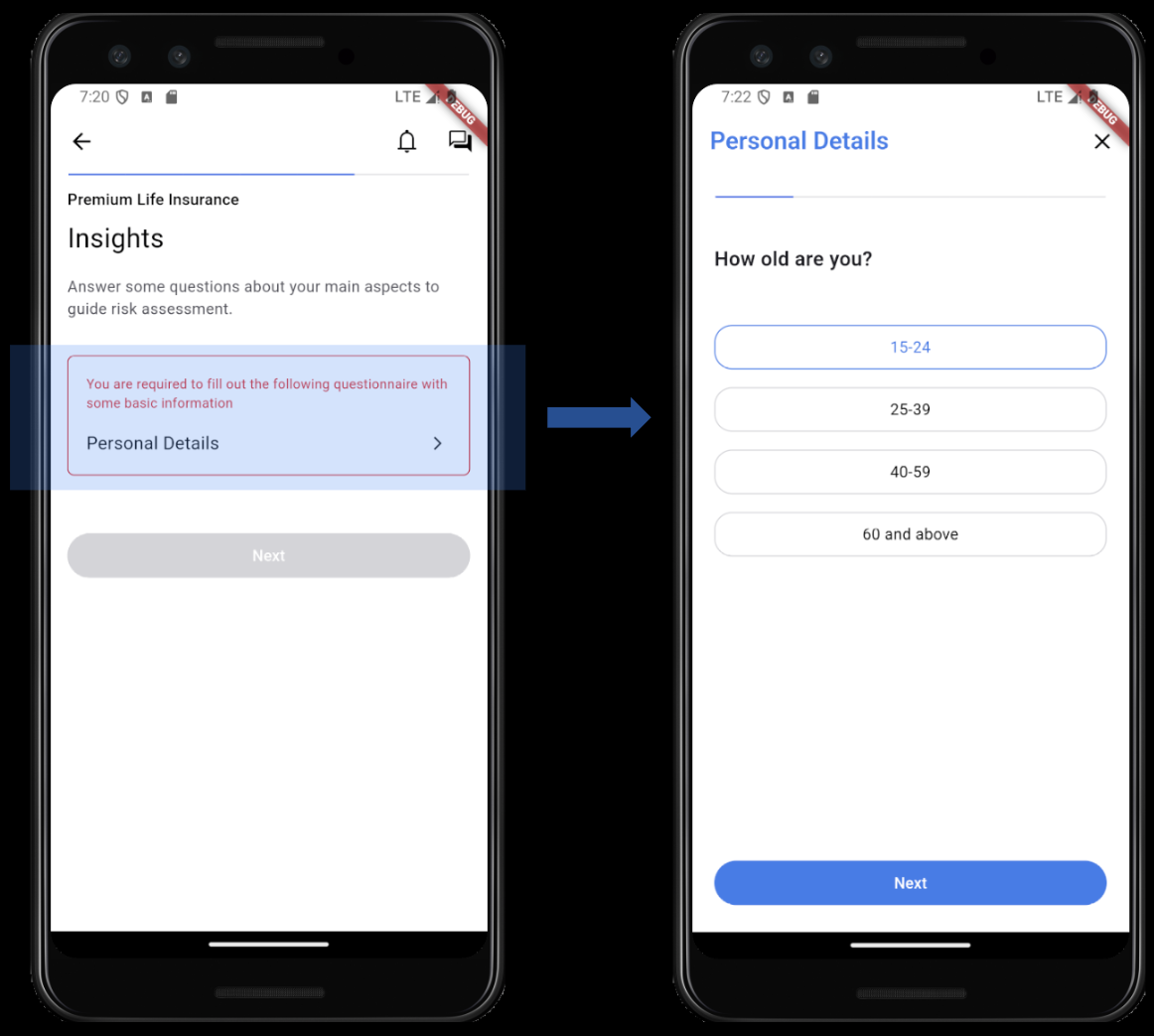}
    \caption{Traditional "Insights" section.}
    \label{fig:traditional_insights}
\end{figure}

Regarding the implementation of the features in the insurance mobile app, the traditional approach begins by simply asking the user about some basic personal details, as visible in Figure~\ref{fig:traditional_insights}.

\begin{figure}[!h]
    \centering
    \includegraphics[width=0.2\textwidth]{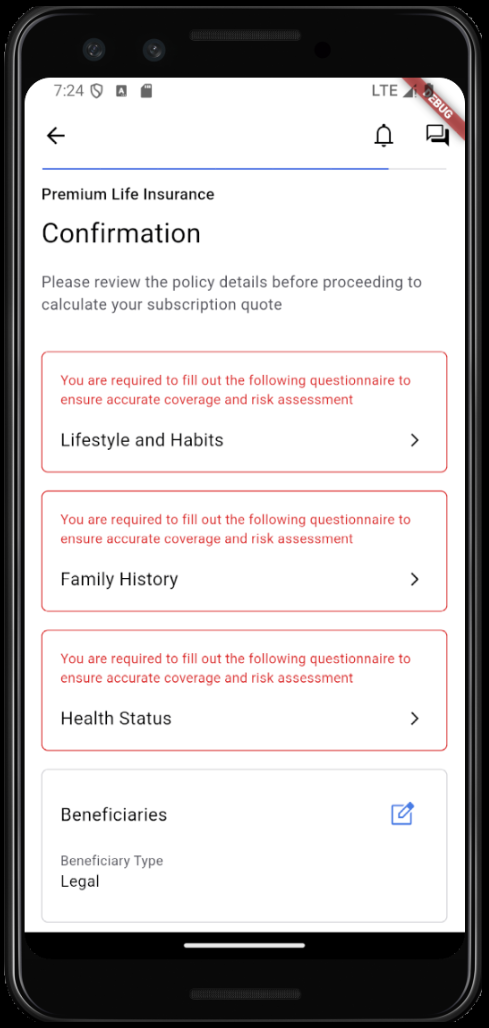}
    \caption{Traditional risk forms.}
    \label{fig:traditional_forms}
\end{figure}

Subsequently, the user is required to respond to three separate sections targeting mortality risk factors in different domains: "Lifestyle \& Habits", "Family History", and "Health Status". Figure~\ref{fig:traditional_forms} shows the interface containing these components. They are composed of pre-defined questions, following the typical traditional subscription.

\begin{figure}[!h]
    \centering
    \includegraphics[width=0.48\textwidth]{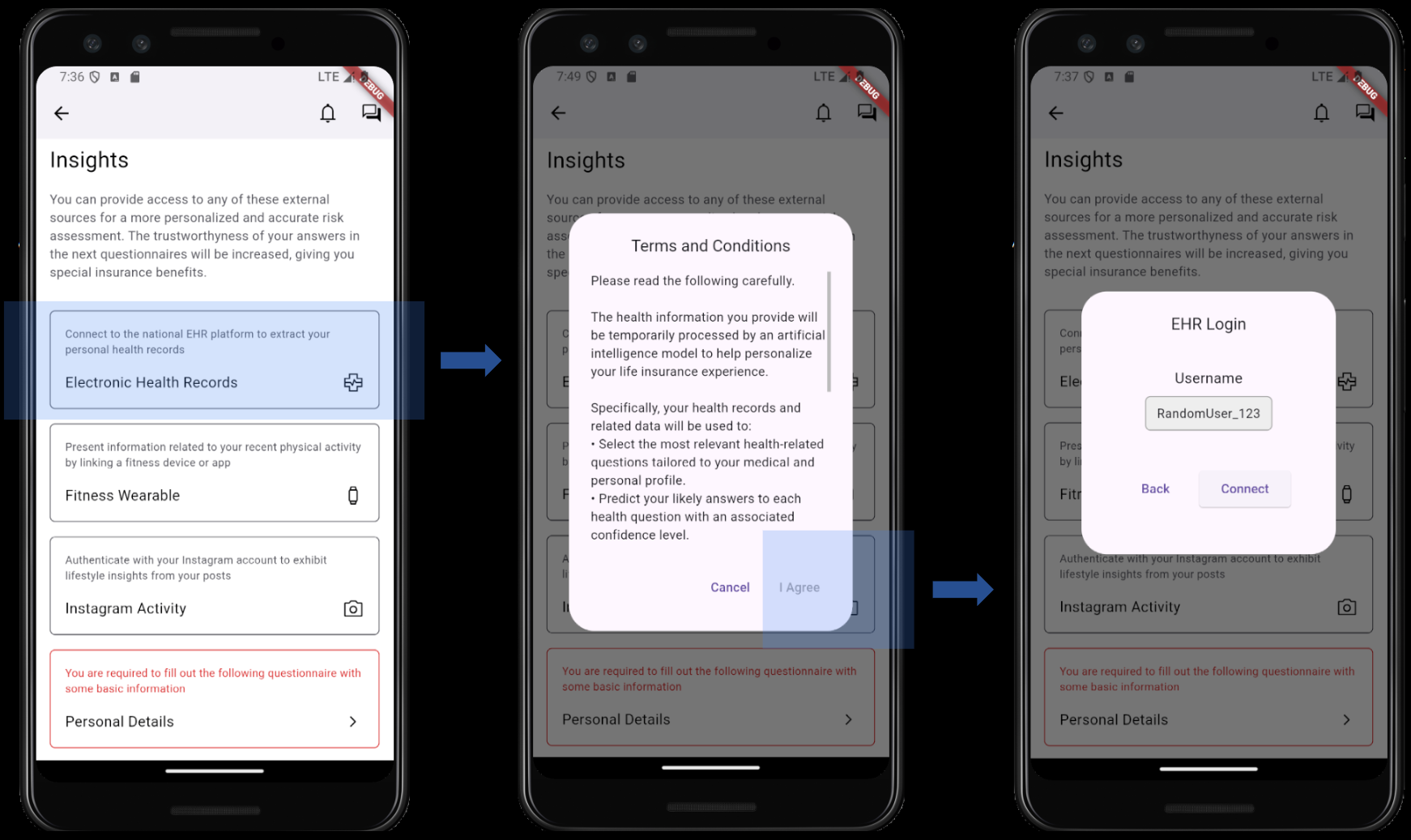}
    \caption{Dynamic "Insights" section.}
    \label{fig:dynamic_insights}
\end{figure}

Unlike the traditional flow, the dynamic approach lets the applicant link external sources (health records, fitness trackers, Instagram posts) to the app. Upon selection of one of them, the user just needs to read and accept the terms and conditions and authenticate with the corresponding platform, as shown in Figure~\ref{fig:dynamic_insights}. In addition to the alternative sources, the user provides personal details, such as age and location.

\begin{figure}[!h]
    \centering
    \includegraphics[width=0.48\textwidth]{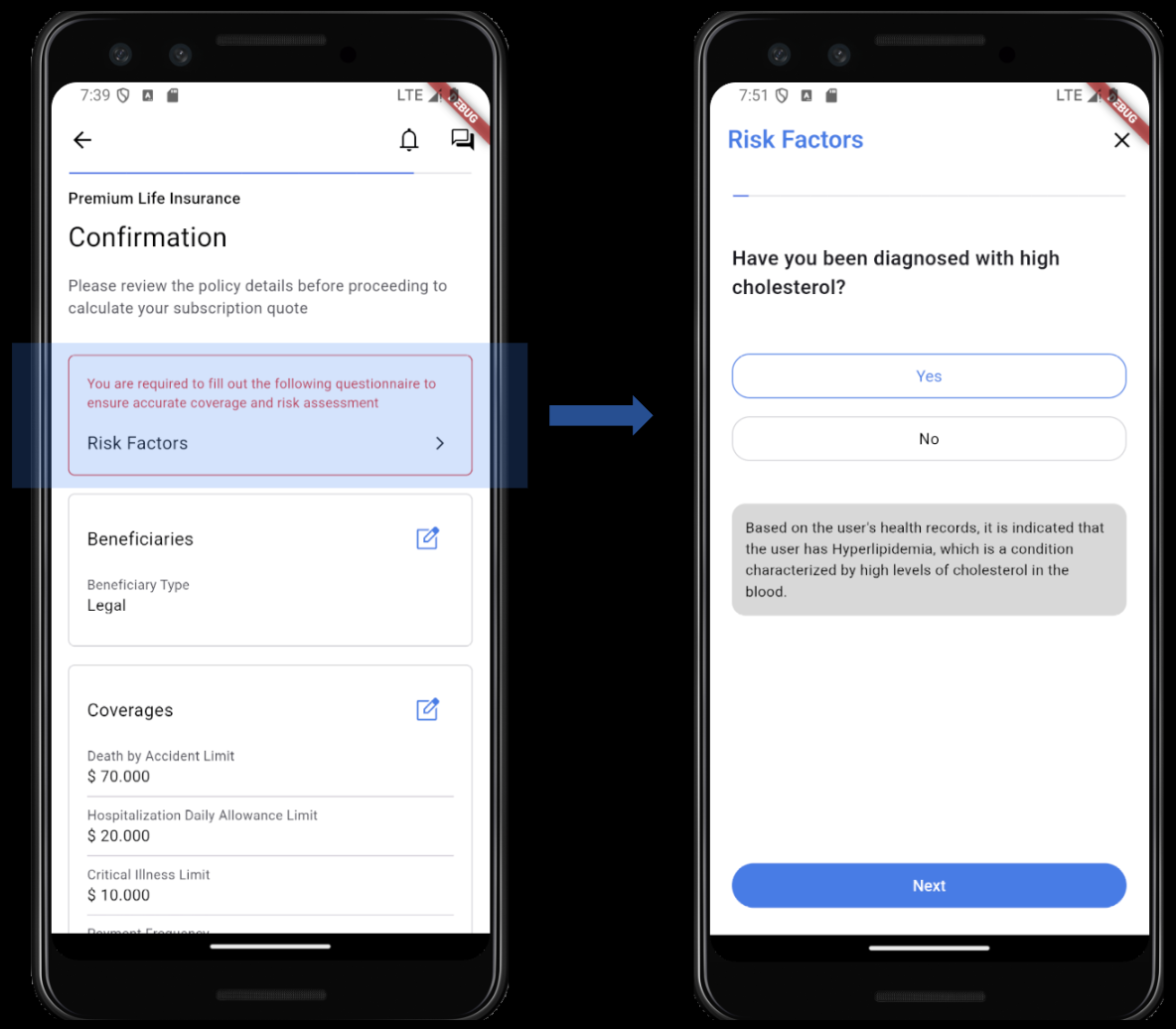}
    \caption{Dynamic risk form.}
    \label{fig:dynamic_forms}
\end{figure}

Following the steps described above, the user is presented with a single dynamic risk form, displayed in Figure~\ref{fig:dynamic_forms}. This questionnaire can include pre-filled questions (with predicted answers and explanations) from the risks identified by the model based on the initial insights, so that the user reviews (and potentially corrects) the analysis performed in a straightforward way. The user is also required to respond to the questions eventually selected by the model for additional feedback.

\subsection{Risk Scoring} \label{sub:arquest/risk_scoring}

There are multiple ways to determine the overall risk score of an applicant based on the questionnaire content. On the one hand, a probabilistic AI model can achieve accurate results through training with real-world trends, but there are interpretability issues due to its black-box nature, with results potentially being influenced by hidden patterns. On the other hand, a deterministic model allows for total control over all variables, even if it increases the difficulty of reaching global optimization.

The priority here is to have a solid algorithm for comparison purposes that is easy to analyze. Therefore, the system resorts to a monotonic additive risk model, where risk starts at zero and can only go up, increasing with non-optimal answers. Moreover, some combinations of responses between related factors increase risk by an additional value.

\section{Evaluation} \label{sec:evaluation}

Two major experiments were conducted to evaluate the solution. The first used the synthetic applicant dataset to compare traditional questionnaires with adaptive versions powered by GPT-3.5 Turbo and GPT-4.1. The second relied on the participation of a small sample of real users to collect user experience insights and overall impressions.

\subsection{Synthetic User Evaluation} \label{sub:evaluation/synthetic_user_evaluation}

Each of the 85 synthetic applicants performed a single traditional subscription and two dynamic subscriptions, one using the GPT-3.5 Turbo model and another with GPT-4.1, to assess whether the older LLM is up to the challenge compared to the newer, more capable one.

Furthermore, these artificial users are previously assigned with "true" risk scores, calculated from the sum of the weights of the answers to all possible questions from the risk dataset, not just the ones that appear in the traditional or dynamic questionnaires.

\begin{figure}[!h]
    \centering
    \includegraphics[width=0.48\textwidth]{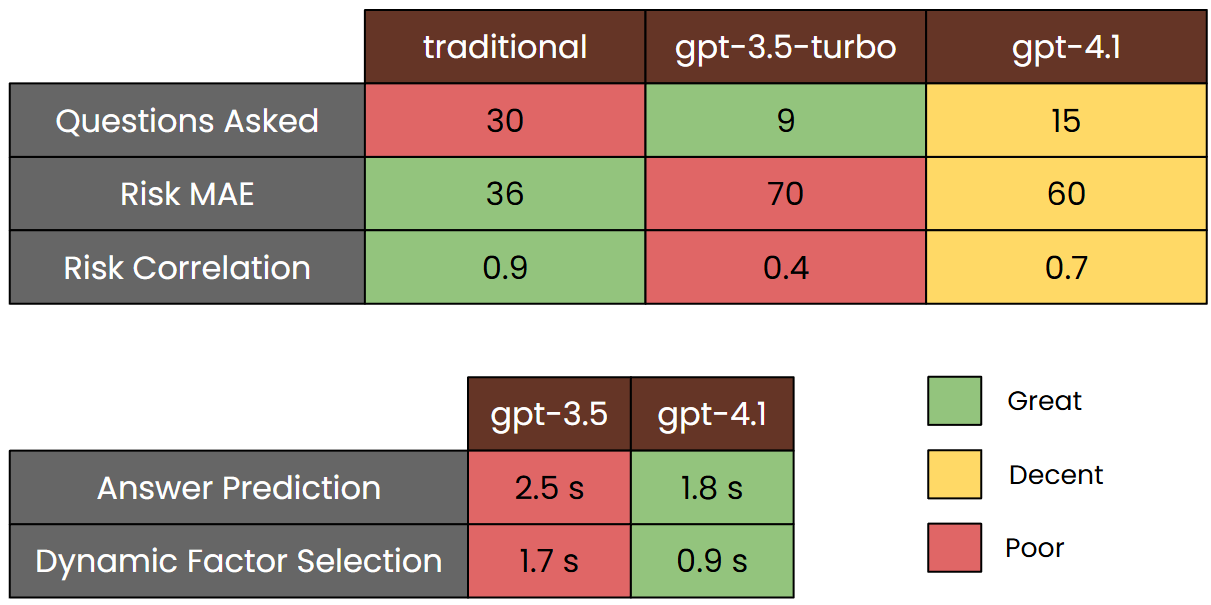}
    \caption{Comparison of approaches for the main evaluation metrics.}
    \label{fig:synthetic_metrics}
\end{figure}

Figure~\ref{fig:synthetic_metrics} showcases how the traditional and dynamic (GPT-3.5 Turbo and GPT-4.1) approaches performed using some relevant metrics. On average, an applicant needs to answer just half the questions of the traditional forms with the dynamic flow, and even less when using GPT-3.5 Turbo. Based on empirical observation, the older LLM seems to be more conservative and cautious, frequently exiting the dynamic questioning process earlier because it interprets ambiguous criteria more narrowly or hesitates to act without strong justification.

However, the risk scores obtained with the traditional approach have a lower mean average error (MAE) or distance to the true risk values, followed by the dynamic method with GPT-4.1 and finally with GPT-3.5 Turbo. The same is observed when analyzing the Pearson correlation with the true scores. Concerning task times, the GPT-4.1 model is significantly faster in both answer prediction and dynamic factor selection.

In general, the dynamic approach using the best model (GPT-4.1) appears to be worse at capturing relevant risk factors than the traditional one, typically by 10-30\% regarding the risk scores obtained. This is mainly due to the significant lack of family history questions within dynamic questionnaires, which is a domain where many users have some risk-related issues.

\subsection{Real User Evaluation} \label{sub:evaluation/real_user_evaluation}

Ten real participants tested both approaches using the insurance mobile app. This sample represents the characteristics of the population to some extent, such as the age and gender distributions. To allow for valid comparisons, half of the users tested the traditional method first, while the other half began with the dynamic flow.

Each validation session comprises the following stages:
\begin{enumerate}
    \item \textbf{Experiment context:} the participant is initially contextualized, with some general instructions.
    \item \textbf{Prior data uploading:} external insights are uploaded beforehand in case the user intends to share them during the experiment.
    \item \textbf{Initial navigation:} the user navigates to the life insurance subscription section inside the mobile app.
    \item \textbf{Testing steps:} the participant tests both the traditional and dynamic approaches, with minimal interference from the orchestrator to avoid influencing perceptions.
    \item \textbf{Final feedback form:} the participant fills a final form that collects user experience feedback.
\end{enumerate}

\begin{figure}[!h]
    \centering
    \includegraphics[width=0.48\textwidth]{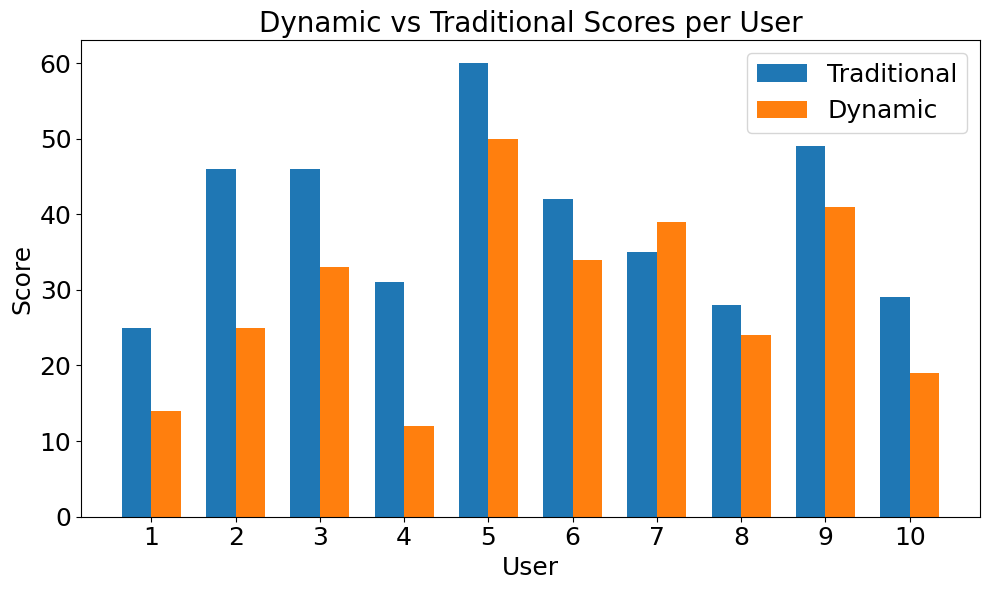}
    \caption{Risk scores per participant using both approaches.}
    \label{fig:participant_scores}
\end{figure}

The risk scores for the participants using both approaches are detailed in Figure~\ref{fig:participant_scores}. Once again, traditional questionnaires gather a higher number of risk factors, leading to higher individual scores. Even so, the values attained using the dynamic forms do not fall far behind, and the relative risk distribution is similar as well. Additionally, the increased trustworthiness that comes with the ARQuest-based methodology enhances the reliability of these results.

Overall, the feedback results indicate a clear preference for the dynamic approach over the traditional static questionnaire. The participants found it more engaging and better suited to their individual context, reinforcing the value of adaptive interactions in insurance risk assessment.

\begin{figure}[!h]
    \centering
    \includegraphics[width=0.48\textwidth]{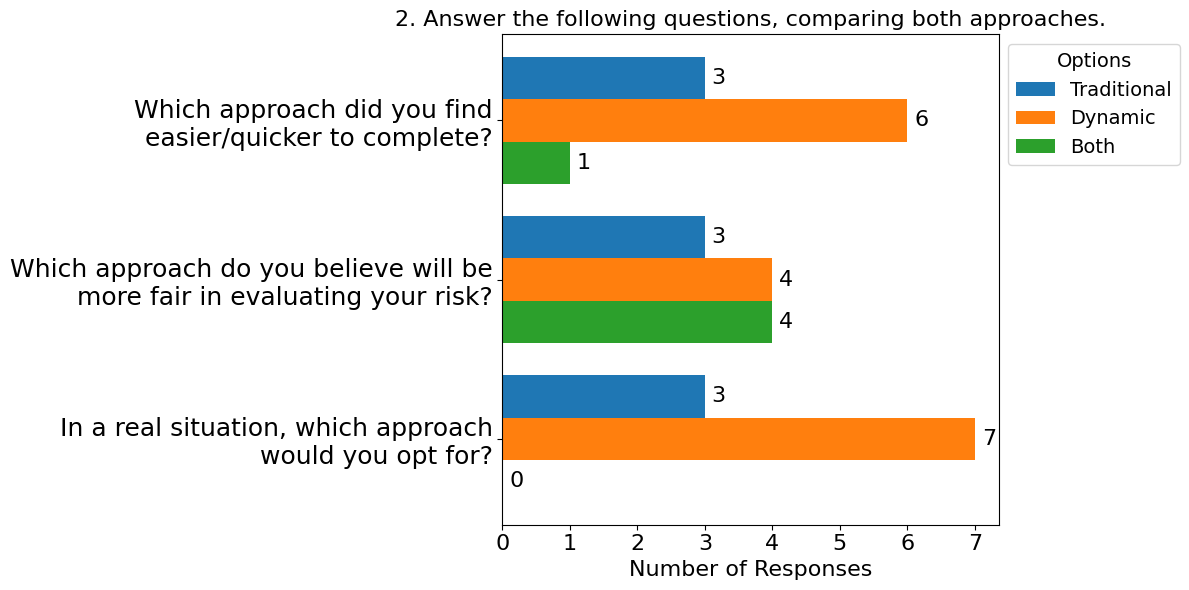}
    \caption{Participant responses for Question 2 of the feedback form.}
    \label{fig:ux_2}
\end{figure}

Most of the participants consider the dynamic approach to be more straightforward and less tedious, with 70\% preferring this solution entirely, visible by the orange bars in Figure~\ref{fig:ux_2}. There is also a general impression that both methods should be similar in determining individual mortality risk, which holds true to some extent.

\begin{figure}[!h]
    \centering
    \includegraphics[width=0.48\textwidth]{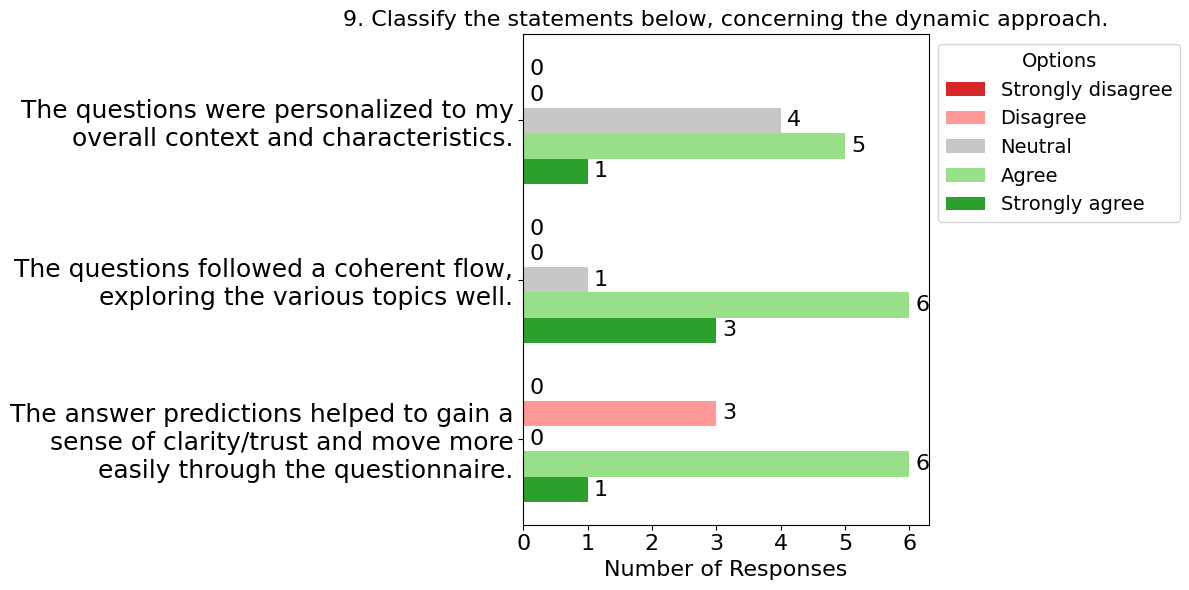}
    \caption{Participant responses for Question 9 of the feedback form.}
    \label{fig:ux_9}
\end{figure}

Focusing on the dynamic approach, participants recognize the effective personalization of the questions presented and the coherent exploration of relevant topics, as elucidated in Figure~\ref{fig:ux_9}. The answer predictions and explanations helped to improve the clarity of the process, with a few exceptions from not-so-satisfied users.

\subsection{Discussion} \label{sub:evaluation/discussion}

After the presented methodologies and analysis, the research questions can be properly answered.

\begin{description}
    \item[\textbf{RQ1}] \textbf{How can AI generate personalized insurance questionnaires in a clear way?}\newline
    By following the ARQuest framework. Briefly, an AI-powered system can simply follow these steps:
    \begin{enumerate}
        \item Analyze insights about the applicant
        \item Identify risks from a pre-made list
        \item Predict those answers beforehand, automating questionnaire completion.
        \item Address additional factors, questioning them one by one.
    \end{enumerate}
    This is a straightforward solution that is easily understandable by both customers and insurers.
    \item[\textbf{RQ2}] \textbf{What concerns arise when integrating external user insights for insurance?}\newline
    Mainly data processing and ethical concerns. Alternative sources are usually composed of unstructured data, which can be substantially contextual and ambiguous. The decisions behind the transformation processes must be thought of carefully to allow easier LLM interpretability. The system should also ensure transparency to build a trusting relationship with the customer and approach each applicant in a way that guarantees fair decision-making.
    \item[\textbf{RQ3}] \textbf{What limitations do current AI models face in capturing individual risk factors?}\newline   Considering the involvement of LLMs in this context, there is a major trade-off between prompt length and model interpretability. Providing the model with the entire content of all questions from the dataset to choose from slows down responses and increases the likelihood of hallucinations occurring. Alternatively, simply listing the names of all risk factors for selection causes a significant loss of nuance within the prompt.
    \item[\textbf{RQ4}] \textbf{How do model confidence and explainability affect AI risk assessments?}\newline
    By improving comprehension of the risk analysis performed by the AI agent for both applicants and insurers, thereby boosting trustworthiness. The confidence level of the predictions is important to identify questions where a high-confidence prediction is drastically altered by the user. The explanations increase transparency and are crucial in proof-checking risk detection, following a case by case analysis.
    \item[\textbf{RQ5}] \textbf{How can personalized underwriting solutions scale to different types of insurance?}\newline
    The ARQuest framework demonstrates how this can be achieved. For a given insurance line, the system needs a knowledge base with the following components: contextually relevant geographical indicators, a quality set of risks and associated questions, a model that calculates a risk score based on the provided user answers, and word embeddings that help detect discrepancies between model guesses and user corrections. 
\end{description}

\section{Conclusion} \label{sec:conclusion}

This study advances personalized questionnaires for insurance underwriting, aiming to improve risk profiling across multiple domains. It addresses a gap in existing research, where the questioning process is often overlooked and few solutions adapt seamlessly across insurance sectors. While alternative data sources are increasingly leveraged, risk assessment is frequently delegated to pre-trained language models capable of capturing complex patterns. Given the sensitivity of insurance, such solutions must also ensure transparency, fairness, and regulatory compliance.

The proposed ARQuest framework personalizes risk assessment by integrating external user insights to identify relevant factors, reducing manual input. It dynamically selects the most informative questions to complete the risk profile and incorporates mechanisms to detect inconsistencies between model reasoning and user responses, strengthening automation reliability.

A life insurance prototype was implemented and evaluated through two experiments: synthetic profiles combining geographic, health, fitness, and social media inputs, and real-user testing via a mobile application. While traditional questionnaires marginally outperformed the adaptive version in detecting certain risk factors, users consistently preferred the dynamic approach for its efficiency, contextual relevance, and improved experience.

Adopting ARQuest offers insurers tangible operational and commercial advantages. Adaptive questioning, coupled with external data integration, shortens application and underwriting times, reduces manual processing costs, and enhances customer satisfaction through a smoother, more personalized process. By capturing accurate data in fewer steps, insurers can improve predictive accuracy, streamline workflows, and refine pricing strategies, strengthening their competitive position.

The main contributions of this work are:

\begin{itemize}
   \item The design of ARQuest, an adaptive questionnaire framework tailored to individual applicant characteristics.

   \item The validation of this framework with both synthetic and real users, demonstrating technical effectiveness, user acceptance, and commercial viability.
\end{itemize}

\subsection{Future Work} \label{sub:conclusion/future_work}

Future enhancements to the ARQuest framework will focus on four main directions. First, the adoption of next-generation LLMs with larger context windows and improved reasoning capabilities is expected to enable more accurate detection of subtle and complex risk patterns. Second, the framework should be tested across multiple insurance lines — such as health, property, and auto — to evaluate its adaptability to diverse risk domains and varying regulatory requirements. Third, fine-tuning the underlying LLM on high-quality, domain-specific datasets could improve both consistency and predictive accuracy, reducing reliance on prompt engineering alone. Finally, integrating agentic AI capabilities capable of autonomously interacting with external tools and data sources would support full end-to-end automation of the underwriting process, further reducing operational overhead and increasing scalability.

\bibliographystyle{ACM-Reference-Format}

\end{document}